\documentclass{article}

\usepackage[nonatbib, final]{nips_2017}

\usepackage[utf8]{inputenc} % allow utf-8 input
\usepackage[T1]{fontenc}    % use 8-bit T1 fonts
\usepackage{hyperref}       % hyperlinks
\usepackage{url}            % simple URL typesetting
\usepackage{booktabs}       % professional-quality tables
\usepackage{amsfonts}       % blackboard math symbols
\usepackage{nicefrac}       % compact symbols for 1/2, etc.
\usepackage{microtype}      % microtypography
\usepackage[pdftex]{graphicx}
\usepackage{listings}
\usepackage{amsmath}
\usepackage{multirow}
\usepackage{amsthm}
\usepackage{ulem}
\usepackage{color}

\title{Semi-supervised learning of hierarchical representations of molecules using neural message passing}

\author{
  Hai Nguyen\thanks{This work was done when the author was an intern of Preferred Networks, Inc.} \\
  Kyoto University \\
  \texttt{hai@kuicr.kyoto-u.ac.jp}
  \And
  Shin-ichi Maeda \\
  Preferred Networks, Inc. \\
  \texttt{ichi@preferred.jp} \\
  \AND
  Kenta Oono \\
  Preferred Network, Inc. \\
  \texttt{oono@preferred.jp}\\
}

\begin{document}

\maketitle

\begin{abstract}
With the rapid increase of compound databases available in medicinal and material science, there is a growing need for learning representations of molecules in a semi-supervised manner.

In this paper, we propose an unsupervised hierarchical feature extraction algorithm for molecules (or more generally, graph-structured objects with fixed number of types of nodes and edges), which is applicable to both unsupervised and semi-supervised tasks.
Our method extends recently proposed Paragraph Vector algorithm\cite{le2014distributed} and incorporates neural message passing \cite{Gilmer2017NMP} to obtain hierarchical representations of subgraphs.

We applied our method to an unsupervised task and demonstrated that it outperforms existing proposed methods in several benchmark datasets.
We also experimentally showed that semi-supervised tasks enhanced predictive performance compared with supervised ones with labeled molecules only.

\end{abstract}

\section{Introduction}

As public databases of chemical compounds, like PubChem \cite{kim2015pubchem}\cite{wang2016pubchem} or ChEMBL \cite{bento2014chembl}, and private databases owned pharmaceutical companies are developed,
there is growing demand to apply them to improve estimation of molecular characteristics or molecular design in medicinal and material science.

One of the most difficult obstacles to achieve this goal is that it can be almost impossible to collect annotated labels. To predict the effectiveness of drugs for a disease, longitudinal studies of patients would be needed to collect ground truth labels. For rare diseases, just acquiring a database of patients may require its own research project. Although high-throughput screening technologies have been developed and the effects of molecules can be evaluated \textit{in vitro}, there still remains a huge gap between experiments \textit{in vitro} and actual effects on a human body, as we can see from the fact that less than 10 percent of drugs passed from the Phase I of clinical trials to approval between 2006 and 2015 \cite{mullard2016parsing}. Likewise, in material science, although we can calculate chemical characteristics that correspond to ground truth labels with first-principles calculation or molecular dynamics, simulation of many-particle systems is still time-consuming. 
Therefore, semi-supervised learning, in which a vast number of unlabeled samples are incorporated with labeled ones to enhance accuracy of models, will play a key role in the mining of molecules in these areas.

In this paper, we propose a novel extension of the \textit{Paragraph Vector} algorithm\cite{le2014distributed}, a well-known unsupervised representation algorithm for documents, to arbitrary graphs and extend it to semi-supervised learning. There have been several approaches for learning representations of graphs. graph2vec \cite{narayanan2017graph2vec} and PATCHY-SAN \cite{niepert2016learning} are two representatives among them that use the Weisfeiler-Lehman (WL) relabelling algorithm \cite{weisfeiler1968reduction}\cite{shervashidze2011weisfeiler} to enumerate rooted subgraphs. Instead, our algorithm is based on neural message passing \cite{gilmer2017neural} to make representations. 
We implemented our algorithm using Chainer neural network framework \cite{chainer_learningsys2015} and experimentally demonstrated the following: 1) Our unsupervised algorithm for learning graph representation outperforms other previously proposed methods on several benchmark datasets, 2) Its extension to semi-supervised tasks achieves better predictive performance than supervised tasks only using labeled molecules.

\section{Related works}
\subsection{Graph convolution}
A \textit{fingerprint} is a fixed- or variable-length vector of binary or float values that reflects the chemical characteristics of a molecule. It is used in several ways, such as in similarity search of chemical compound databases or data mining with machine learning algorithms.
An Extended-connectivity Fingerprint (ECFP) \cite{rogers2010extended}, which is one of the most widely used fingerprint-construction algorithms, encodes all subgraphs whose radius are smaller than some fixed number with a hash function. It uses the Morgan algorithm \cite{morgan1965generation} to enumerate subgraphs of a graph.

Recently, many learning-based fingerprint algorithms have been proposed. \textit{Graph convolutions}, which are extensions of convolution operation from multi-dimensional arrays like images or texts to arbitrary graphs, are attracting much attention. Roughly speaking, they are divided into two types: the message passing neural networks (MPNN) \cite{gilmer2017neural} approach and the spectral approach \cite{defferrard2016convolutional}. Our algorithm is inspired by MPNN. Gilmer et al. showed in \cite{gilmer2017neural} that several MPNN graph convolution algorithms, including neural fingerprints (NFP) \cite{duvenaud2015convolutional} and Gated Graph Neural Networks (GG-NN) \cite{li2015gated}, can be formulated in a unified manner with \textit{message}, \textit{update} and \textit{readout} functions. Note that NFP can be considered as the "soft" version of ECFP. As these models consist of differentiable operations, we can train them with backpropagation.

\subsection{Paragraph vector and its extension}
Representation learning for graphs mainly has dealt with supervised learning, but recently, several researchers have proposed algorithms that learn graph representations in an unsupervised manner.
\textit{Continuous Skip-gram model} \cite{mikolov2013efficient} is an unsupervised algorithm that learns a vector representation of a word. Models are trained so that the representation of a word can predict words that surround it. Specifically, let $W$ be a finite set of distinct words and $D = (w_1, \cdots, w_{|D|})$ be a document where $w_d \in W$ and $|\cdot|$ is the cardinality of a multiset. We write the representation of a word $w\in W$ as $v_w\in \mathbb{R}^d$ where $d$ is some fixed integer. The objective function of the continuous Skip-gram model can be written as follows:
\begin{equation} \label{eq:word2vec}
\sum_{i=1}^{|D|} \sum_{-c\leq j \leq c, j\not = 0} \log P(w_{i+j} \mid w_i).
\end{equation}
Here, $P(w'\mid w) = \exp(v_{w'}^T v_w) / \sum_{u\in W} \exp (v_u^T v_w)$ for $w, w' \in W$ and $c$ is a hyper-parameter that determines the window size. $v^T$ denotes a transpose of a vector $v$ (we use column vectors throughout this paper).
As the computation of eq.\eqref{eq:word2vec} is intractable, Mikolov et al. proposed \textit{negative sampling} \cite{mikolov2013distributed} to change the objective function as
\begin{equation*} \label{eq:ns}
\sum_{i=1}^{|D|} \sum_{-c\leq j \leq c, j \not = 0} \left( \log \sigma (v_{w_{i+j}}^T v_{w_i}) + k \mathbb{E}_{w'\sim P_n} \left[ \log \sigma(-v_{w'}^T v_{w_{i}})  \right] \right)
\end{equation*}
where $\sigma(\cdot)$ is a sigmoid function $\sigma(x) = 1/(1+\exp(-x))$, $k$ is a positive integer, and $P_n$ is some distribution over $W$ called noise distribution. One typical example of the noise distribution is a uniform distribution.
This objective function can be interpreted as a Noise Contrastive Estimation (NCE) \cite{Gutman2012NCE}, as indicated in \cite{mikolov2013distributed}.

The \textit{Paragraph Vector algorithm} \cite{le2014distributed} is an extension of the continuous Skip-gram model that predicts representations of words from that of a document containing them. Formally, we are given a set of documents $\mathcal{D} = \{D_1, \cdots, D_{|\mathcal{D}|}\}$. The $i$-th document $D_i$ is composed of $|D_i|$ words: $D_i = (w_1^i, \cdots, w_{|D_i|}^i)$ where $w_n^i \in W$. We associate a representation $v_D\in \mathbb{R}^d$ for each document $D\in \mathcal{D}$ and $v_w \in \mathbb{R}^d$ for each word $w\in W$. $d$ is again some fixed integer. The model is trained so as to maximize the log-likelihood:
\begin{equation*}
\sum_{i=1}^{|\mathcal{D}|} \sum_{n=1}^{|D_i|}\log P(v_{w_n^i} \mid v_{D_i}),
\end{equation*}
where $P(v_{w} \mid v_{D}) = \exp(v_{w}^T v_D) / \sum_{u\in W} \exp (v_u^T v_D)$.
We can apply negative sampling to this objective function in the same way as the continuous Skip-gram model.

Narayanan et al. extends Paragraph Vector to arbitrary graphs and termed the model \textit{graph2vec} \cite{narayanan2017graph2vec}. Intuitively, a graph and root subgraphs in it for graph2vec correspond to a document and words in Paragraph Vector, respectively. One of the technical contributions of the paper is using the Weisfeiler-Lehman relabelling algorithm \cite{weisfeiler1968reduction} \cite{shervashidze2011weisfeiler} to enumerate all rooted subgraphs up to some specified depth. Our algorithm also can be interpreted as an extension of Paragraph Vector, but instead of listing up rooted subgraphs explicitly, we make use of the neural message passing algorithm recursively to  obtain multi-resolution representations of subgraphs.

\section{Proposed method}
In this section, we present our method for the first contribution to learn hierarchical substructure representations of molecules that can be obtained in an unsupervised learning setting and then present the method to utilize the substructure representation for classification which can be done in a semi-supervised learning setting.

\subsection{Hierarchical substructure representation learning} 
There are hierarchical correlations in each molecule, that is, the atoms tend to connect specific atoms with certain bond, and the neighbors of atoms tend to form specific groups called substructures, the substructures tend to form much larger specific substructures, and so on. 
Note this kind of hierarchical correlations are widely observed in the other 
domains, and our feature extraction method can be applicable to general graph-mining tasks.
Such a correlation for each level, in turn, implies that there exist compact representations of the substructure that  characterize the molecule, and such a compact representation or feature, is often beneficial for the supervised task, especially when the size of the training data is small.
To obtain such a feature for each hierarchical level, we utilize negative sampling \cite{Gutman2012NCE}\cite{mikolov2013distributed} which optimizes the feature by solving a classification task.

Let us denote $h^{l}_{v} \in \mathbb{R}^d$ a discriminative feature vector at level $l$ that can be calculated using the information around the atom $v \in V_m$ of molecule $m \in M$ where $M$ is a given molecule dataset and $V_m$ is a set of all atoms in molecule $m$. We assume that the feature $h^{l}_{v}$ correlates with the molecule vector $u_{m} \in \mathbb{R}^d$ only when the substructure corresponding to $h^{l}_{v}$ is included in
 the molecule $m$ (we denote the case as $C=1$), and decorrelates with $u_{m}$ otherwise ($C=0$). The loss function to obtain the feature is given as follows
\begin{eqnarray}
\sum _{m \in M} \sum_{l=1}^L  \sum_{v \in V_{m}}
\left(
\log p(C=1|h_v^{l}, u_{m}) +
k {\mathbb{E}}_{h_{v'}^{l}\sim p_v^l} \left[\log p(C=0|h_{v'}^{l}, u_{m}) \right] \right) \label{eq:discriminative loss1}
\end{eqnarray}
where $\mathbb{E}_{h_{v'}^{l}\sim p_v^l} [\cdot]$ denotes an expected value with respect to the negative sampler that samples the substructure feature $h_{v'}^{l}$ of molecule $m'$. 
The molecule $m'$ is uniformly sampled from the given molecule dataset $M$, and $h^{l}_{v'}$ is computed for randomly chosen atom $v'$ at level $l$ unless it matches the substructure $h^{l}_{v}$. If the randomly chosen substructure $h^{l}_{v'}$ matches $h^{l}_{v}$, then the molecule $m'$ is rejected and another molecule is resampled and this procedure is repeated until the molecule $m'$ is accepted.
$k$ denotes a positive scalar that determines the number of samples taken by the negative sampler $p_v^l$. In the experiment, we set $k=10$. 
We define the model $p(C=1|h_v^{l}, u_{m})$ as
\begin{eqnarray}
p(C=1|h_v^{l}, u_{m}) = \sigma( u_{m}^T h_v^{l}). \label{eq:sigmoid}
\end{eqnarray}
By substituting eq.\eqref{eq:sigmoid} to eq.\eqref{eq:discriminative loss1}, the loss function becomes
\begin{eqnarray}
\sum _{m \in M} \sum_{l=1}^L  \sum_{v \in V_{m}}
\left( 
\log \sigma(\gamma u_{m}^T h_v^{l}) +
k {\mathbb{E}}_{h_{v'}^{l}\sim p_v^l} \left[\log \sigma(-\gamma u_{m}^T h_{v'}^{l}) \right]
\right). \label{eq:discriminative loss2}
\end{eqnarray}
We maximize the above objective function with respect to the parameters of $h_v^{l}$ which is explained later, and all of the molecule vectors $u_{m}$  $(m \in M)$ directly.
During the training, the expectation $\mathbb{E}_{p_v^l} [\cdot]$ is replaced by $k$-times sampling from $p_v^l$.  
%Note that $p(C=0|h_v^{l}, h_{mol}) = \sigma(-\gamma h_{mol}^T h_v^{l}) $

The state $h^{l}_v$ is computed by the neural network in an hierarchical manner.
At each level $l$, the computation follows the neural message passing \cite{Gilmer2017NMP}.

\begin{eqnarray}
    m^{l+1}_{v} = \sum_{w \in N(v)}H_{e(v, w)}h^{l}_{w} \label{eq: message_passing} \\
     h^{l+1}_{v} = \mathbf{ \sigma} (h^{l}_{v} + m^{l+1}_{v})\label{eq: update}
\end{eqnarray}
where $e(v, w)$ indicates the type of the bond between two atoms $v$ and $w$ i.e., one of four types of bond in our implementation: single, double, triple and aromatic, and $N(v)$ denotes the set of neighborhood atoms of atom $v$. $H_{e(v,w)}$ is a $d \times d$ matrix, and $\mathbf{ \sigma}$ is an element-wise sigmoid function (with a slight abuse of notation).
Note that $H_{e(v,w)}$ only depends on the type of the bond, and 
 it is shared for any atom and molecule so that it can compute the substructure representation for any size and atomic composition of molecules.
Because the all variables are differentiable with respect to the parameters of neural networks, we can optimize the loss function by any kind of stochastic gradient descent.

\begin{figure}
    \centering
    \includegraphics[width=1.0\textwidth ]{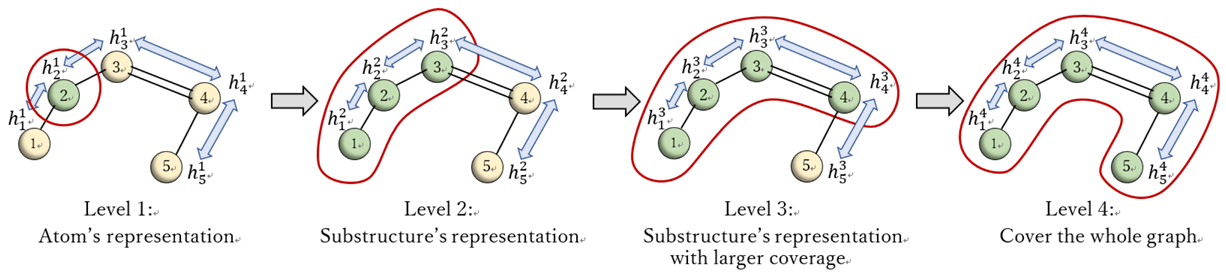} %{mess_update.png}
    \caption{Message passing and update mechanism used to represent multi-resolution substructures.
    Each circle represents an atom, and the edge between circles represents the connection between adjacent atoms. The atoms used to compute the substructure representation around atom 2 are shaded.
}
    \label{fig:message-passing-update}
\end{figure}

As can be seen from the eqs.\eqref{eq: message_passing} and \eqref{eq: update}, neural message passing consists of two functions, message function and update function. The message  function applied on atom $v$ is to collect information from neighbors and the update function is to update the feature at $v$ based on the collected information and the former feature of $v$ as in \eqref{eq: message_passing}. By applying \eqref{eq: message_passing} and \eqref{eq: update} several times, the updated feature at atom $v$ can be used to represent substructures with root of atom $v$ as illustrated in Figure \ref{fig:message-passing-update}.

\begin{figure}
    \centering
    \includegraphics[width=0.4\textwidth ]{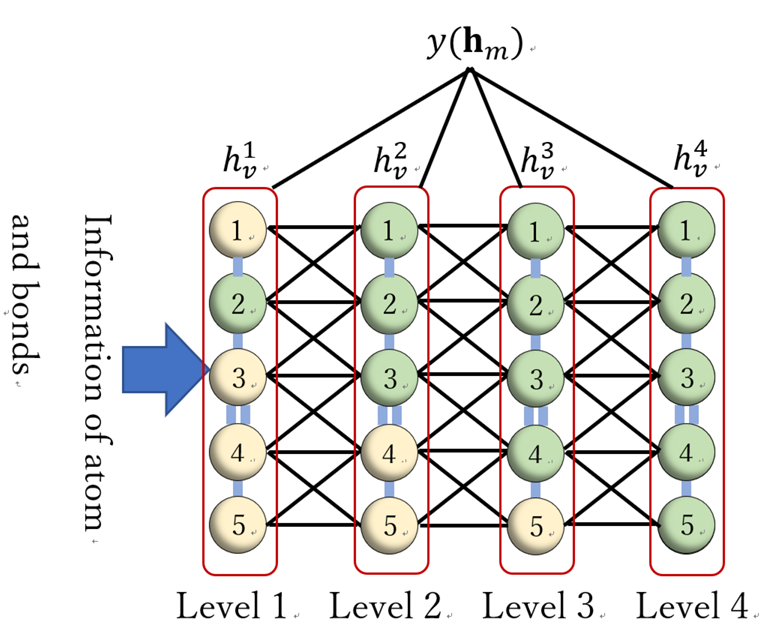} %{mess_update.png}
    \caption{Overall architecture.
    Every multi-resolution substructure computed from the input graph contributes to the computation of the output $y$.}
    \label{fig:architecture}
\end{figure}

\subsection{Classification using substructure representation}
After obtaining the representation of rooted substructures and molecules, either of them can be used for classification. However, we argue that directly using molecule representations is not effective because they lose too much 
information of the molecule to predict the properties of the molecule,
suggesting that using the set of features of rooted substructures should be better.
%However, we argue that directly using molecular representation for classification is not effective because they do not contain discriminative information for classification task, suggesting that substructure representation should be utilized instead.

Our method of using substructures for classification is motivated by neural fingerprint (NFP) proposed in \cite{duvenaud2015convolutional}. More specifically, a given molecule $m$ is composed of substructures with $L$ different levels (also including atoms at 1st level). 
We construct the following readout function for the classification:
\begin{equation}
% out = {\rm{NN}} \left(\sum^{L}_{l=1}\sum^{n_{m}}_{i=1}f(W h^{l}_{i}) \right)
y({\bf{h}}_m) = {\rm{NN}} \left(\sum^{L}_{l=1}\sum_{v\in V_m}f(W h^{l}_{v}) \right)
\end{equation}
where ${\bf{h}}_m$ is a set of features computed from molecule $m$ and $y({\bf{h}}_m)$ is output of the classifier. 
$W$ is a weight matrix and commonly used by all substructures, $f$ is  non-linear function and ${\rm{NN}}$ is a neural network to map the input to the output. In our experiments, $f$ is a softmax function and ${\rm{NN}}$ is a two-layer neural network.
The overall architecture is shown in Figure 2.

\section{A semi-supervised framework for prediction of molecular properties}
In this section, we describe our second contribution of this work. This is motivated by the fact that the number of molecules with known properties is sometimes few while the information of the structure of many undiscovered ones are available. In other words, we have a lot of unlabeled samples and a relatively small number of labeled samples. Therefore we address the problem of how to take advantage of unlabeled molecules to improve prediction. 

The main goal of the first contribution we presented above is to introduce an efficient way to learn substructures of molecules in an unsupervised setting. We continue to propose a semi-supervised learning approach for classifying a large number of molecules with unknown properties.

\textbf{Problem setting:} Given a set of labeled  molecules  $\mathcal{M}^{L}=\{m_{1}, \cdots , m_{|\mathcal{M}^{L}|}\}$ with corresponding output $\{o_{1}, \cdots ,o_{|\mathcal{M}^{L}|}\}$, and  a set of undiscovered molecules (or unlabeled samples) $\mathcal{M}^{U}= \{ m_{|\mathcal{M}^{L}| + 1}, \cdots, m_{|\mathcal{M}^{L}| + |\mathcal{M}^{U}|} \}$ where $|\mathcal{M}^{U}| \gg |\mathcal{M}^{L}|$.

We try to minimize the following objective function:
\begin{equation}
%\sum_{m \in M^{L}} {\rm{Loss}}(m, y({\bf{h}}_m), o_m) + \lambda \sum_{m \in M^{U} \cup M^{L}} {\rm{Reg}}(m, {\bf{h}}_m)   \label{eq:objective}
\sum_{i=1}^{|\mathcal{M}^{L}|} {\rm{Loss}}(m_i, y({\bf{h}}_{m_i}), o_i) + \lambda \sum_{j=1}^{|\mathcal{M}^{L}| + |\mathcal{M}^{U}|} {\rm{Reg}}(m_j, {\bf{h}}_{m_j})   \label{eq:objective}
\end{equation}
where ${\rm{Loss}}(m_i, y({\bf{h}}_{m_i}), o_i)$ is defined as the loss function of molecule $m$ that measures the discrepancy of the classifier output $y({\bf{h}}_{m_i})$ and the true output $o_{i}$, and ${\rm{Reg}}(m_j, {\bf{h}}_{m_j})$ is the regularization term put on the molecule $m_j$ to optimize the feature ${\bf{h}}_m$ as defined in eq.~\eqref{eq:discriminative loss2}.
The hyperparameter $\lambda$ controls the relative weight between the purely supervised loss and regularization term. 
As we can guess from the objective function, the features ${\bf{h}}_{m_i}$ are trained so that it can predict the molecular property for the labeled dataset while it keeps rich discriminative information of the molecular for both of the labeled and unlabeled dataset.
Note that the objective function \eqref{eq:objective} can again be optimized by stochastic gradient descent.%backpropagation \cite{rumelhart1988learning}.

\section{Experiments and results}
In this section, we present our experiments and results to validate the effectiveness of our hierarchical substructure representation and semi-supervised approach based on the extracted substructure representation. For the evaluation of the our substructure feature, we train the network in an unsupervised manner to evaluate the genuine effectiveness of the feature. We evaluate the accuracy of our method both in unsupervised and semi-supervised learning tasks corresponding to the two aforementioned contributions. 
Through all the experiments, the dimension of the substructure feature vector $h_v^l$ is set to 100 which is empirically determined.
The hyperparameter $\gamma$ in eq.\eqref{eq:discriminative loss2} was selected among $\gamma \in \{0.1, 0.5 ,1.0\}$, and was set to 0.5 due to good yields on the training dataset.
As for the hyperparameter $\lambda$ used in the loss function \eqref{eq:objective} of the semi-supervised learning, it was set to 0.5 thanks to experimentation. 
For the implementation, we used the neural network framework Chainer \cite{chainer_learningsys2015}. All the experiments were conducted on a Macbook computer with 2.7 GHz Intel Core i5 and 8GB memory. 

\subsection{Unsupervised learning task}
\textbf{Datasets:} We used MUTAG \cite{Debnath1991MUTAG} and PTC \cite{Helma2001} as two benchmark graph classification data sets for our experiments. MUTAG consists of 188 chemical compounds and their class labels indicate whether the compound has a mutagenic effect on a specific bacteria. PTC comprises of 344 compounds and their classes indicate carcinogenicity on rats. Both are binary classification tasks.

\textbf{Comparison:} Our proposed approach is compared with the existing proposed methods, including node2vec \cite{Aditya2016node2vec}, WL kernel \cite{shervashidze2011weisfeiler}, Deep WL kernel \cite{deepwlkernel} and graph2vec \cite{narayanan2017graph2vec}. 
To evaluate the usefulness of the derived features from each unsupervised learning method, we used the strategy from the previous study \cite{narayanan2017graph2vec} of evaluating the performance on the supervised task with the same SVM classifier, but using the features derived from the unsupervised learning.
For fair comparison, the detailed experimental setting also follows the previous study \cite{narayanan2017graph2vec}, i.e., the ratio of the size of training dataset and the test dataset is 9:1, the division of the training dataset and test dataset are at random, and the division and the training is repeated ten times. Then the average accuracy and the standard deviation for the ten trials was evaluated. 
The results on the two dataset obtained by our method are summarized in Table 1, with existing methods reported by graph2vec\cite{narayanan2017graph2vec} for reference.
They show that our method is better than the above one in terms of the predictive performance on the two datasets, which implies our substructure feature is more informative than the others.

\begin{table}[ht]
    \centering
    \begin{tabular}{|c|c|c|c|c|c|c|}
        \hline
        \textbf{Datasets} & \textbf{node2vec} & \textbf{sub2vec} & \textbf{graph2vec} & \textbf{WL kernel} & \textbf{Deep WL} & \textbf{Ours} \\ \hline
        \textbf{MUTAG} & 72.63$\pm$10.2 & 61.05$\pm$15.80 & 83.15$\pm$9.25 & 80.63$\pm$3.07 & 82.95$\pm$1.96 & \textbf{86.46}$\pm$5.97 \\ \hline
        \textbf{PTC} & 58.85$\pm$8.00 & 59.99$\pm$6.38 & 60.17$\pm$6.86 & 59.61$\pm$2.79& 59.04$\pm$1.09 & \textbf{62.86}$\pm$5.71 \\ \hline
    \end{tabular}
    \caption{Comparison of the usefulness of the features obtained by unsupervised learning methods. The usefulness of the features are evaluated by the performance on the supervised task using the same SVM classifier.  The figures of each cell denote the average accuracy and the standard deviation.}%PR: Is it supervised or unspervised? Text says both.
    \label{tab:my_label}
\end{table}

\subsection{Semi-supervised learning task}
\textbf{Datasets:} Two typical data sets, including solubility \cite{delaney2004esol} and drug efficacy \cite{gamo2010thousands} are selected to compare the performance of NFP (supervised learning model) and the proposed semi-supervised approach. The datasets consist of 1144 molecules (solubility) and 10000 molecules (drug efficacy), respectively. In our experiments, we select a relatively small subset of molecules as the labeled set and the rest is unlabeled.

\textbf{Comparison:} Our purpose is to compare the supervised method with a few number of labeled samples and semi-supervised method with additional set of unlabeled samples. Similar to the unsupervised task, training dataset and test dataset are randomly divided with the ratio 9:1 and trained on ten times.  
The results are reported in Table 2. It is evident that our proposed semi-supervised task is better than supervised method with few labeled samples, showing the effectiveness of our semi-supervised learning approach on prediction of molecular properties.

\begin{table}[ht]

\begin{center}
    \begin{tabular}{|c|c|c|c|}
     \hline
     \multirow{2}{*}{\textbf{data sets}} & \multirow{2}{*}{\textbf{labeled/ all}} & \multicolumn{2}{|c|}{\textbf{Methods}} \\ \cline{3-4}
                               &                               & \textbf{NFP (avg $\pm$ std)}  & \textbf{SemiNFP (avg $\pm$ std)} \\ \hline
    \multirow{4}{*}{Solubility (log Mol/L)} & 3\% & 2.26 $\pm$ 0.07 & 1.85 $\pm$ 0.03 \\ \cline{2-4}
                           & 6\% & 1.8 $\pm$ 0.03 & 1.56 $\pm$ 0.07 \\ \cline{2-4}
                           & 12\% & 1.48 $\pm$ 0.2 & 1.24 $\pm$ 0.1 \\ \cline{2-4}
                           & 18\% & 1.21 $\pm$ 0.12 & 1.12 $\pm$ 0.35 \\ \hline
    \multirow{4}{*}{Drug efficacy $EC_{50}$ in nM} & 3\% & 1.74$\pm$0.24 & 1.59 $\pm$ 0.12 \\ \cline{2-4}
                           & 6\% & 1.55$\pm$0.17 & 1.42$\pm$0.32 \\ \cline{2-4}
                           & 12\% & 1.57$\pm$0.11 & 1.41$\pm$0.26 \\ \cline{2-4}
                           & 18\% & 1.51$\pm$0.21 & 1.35$\pm$0.19 \\ \hline
\end{tabular}
\caption{Comparison of supervised and semi-supervised learning tasks with a few labeled molecules}
\end{center}
\end{table}

\section{Conclusion}
In this paper, we propose a novel hierarchical feature extraction method that describes the molecular characteristics in a compact vector form in an unsupervised setting. The features are trained so that it keeps the discriminative feature of molecules as much as possible at each hierarchical level.
This feature extraction method not only yields the state-of-the-art performance on the unsupervised task but also  
makes it possible to introduce a semi-supervised learning framework to the supervised tasks and successfully demonstrates the effectiveness of our semi-supervised learning. 
To the best of our knowledge, this is the first study that brings the semi-supervised learning framework to the prediction task of molecular properties.

\bibliographystyle{plain}
\bibliography{reference}
\end{document}